\documentclass[lettersize,journal]{IEEEtran}
\usepackage{amsmath,amsfonts}
\usepackage{algorithmic}
\usepackage{algorithm}
\usepackage{array}
\usepackage[caption=false,font=normalsize,labelfont=sf,textfont=sf]{subfig}
\usepackage{textcomp}
\usepackage{stfloats}
\usepackage{url}
\usepackage{verbatim}
\usepackage{graphicx}
\usepackage{cite}
\usepackage{bm}
\newtheorem{definition}{Definition}

\usepackage{mathrsfs}  
\usepackage{xspace}
\newcommand{\algo}{CC\@\xspace}

\newcommand{\E}{\mathbb{E}}
\newcommand{\expec}[1]{\mathbb{E}\left[#1\right]}

\usepackage{mathtools}

\newcommand{\argmax}{\operatornamewithlimits{argmax}}

\usepackage[colorinlistoftodos, textwidth=30mm]{todonotes}

\usepackage{arydshln}

\begin{document}

%\title{Configure \& Conquer in Marine Resource Optimization}
\title{Optimizing Empty Container Repositioning and Fleet Deployment via Configurable Semi-POMDPs}

\author{Riccardo Poiani, Ciprian Stirbu, Alberto Maria Metelli, Marcello Restelli
\thanks{R. Poiani, A.M. Metelli and M. Restelli are with the Dipartimento di Elettronica, Informazione e Bioingegneria, Politecnico di Milano, Milan, Italy. R. Poiani started working on this project while he was an intern at InstaDeep, Paris, France. C. Stirbu works for InstaDeep, Paris, France.}
\thanks{This work has been submitted to the IEEE for possible publication. Copyright may be transferred without notice, after which this version may no longer be accessible.}
}

%\author{IEEE Publication Technology,~\IEEEmembership{Staff,~IEEE,}
        % <-this % stops a space
%\thanks{This paper was produced by the IEEE Publication Technology Group. They are in Piscataway, NJ.}% <-this % stops a space
%\thanks{Manuscript received April 19, 2021; revised August 16, 2021.}}



\maketitle

\begin{abstract}
With the continuous growth of the global economy and markets, resource imbalance has risen to be one of the central issues in real logistic scenarios. In marine transportation, this trade imbalance leads to Empty Container Repositioning (ECR) problems. Once the freight has been delivered from an exporting country to an importing one, the laden will turn into empty containers that need to be repositioned to satisfy new goods requests in exporting countries. In such problems, the performance that any cooperative repositioning policy can achieve strictly depends on the routes that vessels will follow (i.e., fleet deployment). Historically, Operation Research (OR) approaches were proposed to jointly optimize the repositioning policy along with the fleet of vessels. However, the stochasticity of future supply and demand of containers, together with black-box and non-linear constraints that are present within the environment, make these approaches unsuitable for these scenarios. In this paper, we introduce a novel framework, Configurable Semi-POMDPs, to model this type of problems. Furthermore, we provide a two-stage learning algorithm, ``Configure \& Conquer" (CC), that first configures the environment by finding an approximation of the optimal fleet deployment strategy, and then ``conquers" it by learning an ECR policy in this tuned environmental setting. We validate our approach in large and real-world instances of the problem. Our experiments highlight that CC avoids the pitfalls of OR methods and that it is successful at optimizing both the ECR policy and the fleet of vessels, leading to superior performance in world trade environments.
\end{abstract}


\section{Introduction}
\IEEEPARstart{N}{owadays} marine transportation is crucial for the world’s economy: $80\%$ of the global trade is carried by sea, and most of the world’s marine cargo is transported in containers \cite{long2012sample,li2019cooperative}. In 2004, over $60\%$ of the total amount of goods shipped by sea were containerized, while some routes among economically strong countries are containerized up to $100\%$ \cite{long2012sample}. When using such methods to convey goods, problems arise due to the joint combination of the inner nature of container flow with the imbalance of global trade between different regions. For instance, once the freight has been delivered from an exporting country to an importing one, the laden will turn into empty containers that need to be repositioned to satisfy new goods requests in exporting countries. This problem takes the name of Empty Container Repositioning (ECR) \cite{song2015empty}. In ECR, when a vessel arrives at a port to discarghe laden, the port has two options: discharge a certain amount of empty containers that are present on the vessel, or load the vessel with empty containers from its stock. The goal is to find \emph{cooperative} repositioning policies that minimize the shortage of demand of empty containers in a given horizon. ECR can be a very costly activity in complex logistic networks and, even if it does not directly generate income, it can account for about $20\%$ of the total costs for shipping companies \cite{song2015empty}.
Thus, building efficient ECR strategies is a crucial point for real-world logistic scenarios.

As highlighted in previous studies \cite{tran2015literature}, the demand of empty containers that can be satisfied in a given horizon by any ECR policy strictly depends on the routes that the vessels of the given shipping company will follow. From an intuitive point of view, this is clear if we consider a scenario with two routes that have no ports in common. Suppose that most of demand of empty containers concerns ports that are present on the first route. If most of the vessels that the shipping company owns follow the second route, any ECR policy will have poor performance since the cooperation ability of the network to exchange containers is limited by the poor assignments of vessels to routes. In the marine transportation literature, the problem of assigning vessel to routes to maximize some given target function takes the name of Fleet Deployment (FD) \cite{tran2015literature}. More specifically, in FD, the set of routes is predetermined by the shipping company, and the task is to select the starting port on a route for each vessel that the shipping company owns. 

In this paper, we study the adoption of FD techniques to improve ECR policies. Historically, Operation Research (OR) approaches were proposed to jointly optimize the repositioning policy along with the fleet of vessels \cite{wang2017container}. However, the stochasticity within the environment together with black-boxed and non-linear constraints that are present in the environment makes them unsuitable for such complex scenarios \cite{li2019cooperative}. Multi-agent Reinforcement Learning (MARL) techniques, on the other hand, have recently achieved success at limiting these problems in ECR settings in which the assignments of vessels to routes is predetermined and given to the learners \cite{li2019cooperative,shi2020cooperative}. They model the problem as a Semi-Partially Observable Markov Decision Process (Semi-POMDP) and propose ad-hoc neural architectures to learn cooperative policies. \footnote{Notice that the ``Semi" component arises from the fact that the ECR problem is intrinsically event-driven: a repositioning action needs to be taken only when a vessel arrives at a given port.} 
%\todoam{"Semi-Partially Observable Markov Decision Process" Si dice così? Oppure "Partially Observable Semi-Markov Decision Process "?}

%\todorp{In shi2020cooperative lo chiamano Semi-POMDP. Di per se e' un setting non studiato ad-hoc formalmente. Io comunque lo terrei cosi perche rende l'acronimo piu comprensibile.}
 
In this paper, our focus is on a more complex problem, that is \emph{jointly optimizing} the repositioning policy (ECR) together with the fleet of vessels (FD). In this sense, we are jointly solving the ECR+FD problem.
More specifically, from an agents' perspective, the fleet of vessels can be seen as features of the environment that can be optimized to reach higher performances. In this sense, for single-agent problems, Configurable Markov Decision Processes (Conf-MDPs) \cite{metelli2018configurable,metelli2019policy,metelli2019reinforcement} have recently been introduced to extend the Markov Decision Process (MDP) \cite{puterman} framework to account for environmental configurations. In Conf-MDPs, an \emph{agent} and a \emph{configurator} are responsible for finding the optimal policy-configuration pair. This is clearly related to our application scenario: our agents are in charge of deciding which container repositioning policy to play, whereas the configurator is entitled to select the fleet of vessels. While the early works \cite{metelli2018configurable,metelli2019policy,metelli2019reinforcement} focused on the case in which agent and configurator share the same objective, in \cite{ramponi2021learning}, the setting has been extended to the case in which the configurator and the agent have different (and, possibly, adversarial) goals. Although these approaches have strong theoretical guarantees, how to successfully scale them to more complex domains remains an open question. Indeed, in a \emph{multi-agent cooperative} setting, the dimension of the problem explodes with the number of agents. Moreover, the intrinsic non-stationarity present in multi-agent systems significantly complicates the learning process \cite{zhang2021multi}. We also note that all the previous methods assume the state to be fully observable; in ECR, however, the agents operates under \emph{partial observability}, that introduces an additional challenge.\footnote{Notice that the optimal repositioning policy is history-dependent.} 

The contributions of our works are summarized as follows:
\begin{itemize}
\item 
%After providing some background (Section \ref{sec:prelim}) and revising the literature (Section \ref{sec:literature}), 
We introduce the \emph{Configurable Semi-POMDPs} (Conf-Semi-POMDPs), a novel framework whose goal is extending Conf-MDPs to the more complex multi-agent, partially-observable dynamics of the ECR+FD problem (Section \ref{sec:setting}). In particular, we focus on the cases in which the configuration of the environment (i.e., assignment of vessels to routes) is decided by a central entity (i.e., \textit{configurator}); that, in our case, is the shipping company. 
\item To solve Conf-Semi-POMDP, we propose a general two-step solution algorithm called ``Configure \& Conquer" (CC) (Section \ref{sec:method}). The goal of \algo is to build solutions that successfully \emph{scale} the joint optimization process (i.e., policy and configurations) to our large multi-agent systems. \algo first optimizes the configurator to output an approximation of the optimal configuration (i.e., fleet deployment) and then ``conquers” it by learning a policy in this tuned environmental setting (i.e., the ECR cooperative policy). The main intuition that \algo exploits in its configure step is that to compare two distinct configurations one can even leverage suboptimal policies. 
\item We validate our approach in large and real-world instances of the ECR+FD problem (Section \ref{sec:exp}). Our experiments show that CC avoids the pitfalls of OR methods and that it is successful at optimizing both the ECR policy and the fleet of vessels, leading to superior performance in world trade environments.
\end{itemize}

%Finally, we remark that our framework and algorithm are \emph{general} and can be applied to a broad range of range of real-world applications. Indeed, in many multi-agent settings the parameters of the environment can be configured so that the multi-agent system can increase its overall performance. For instance, in unmanned aerial vehicles fleets \cite{floriano2019planning}, each agent can be equipped with different communication components, and we might want to find the best equipment for a certain goal subject to some cost metric. In target tracking by a team of sensors \cite{nair2005networked}, metrics of interest such as the number of successful scans are subject to how the sensor network is configured (e.g., which sensor can cover which zones and sensors' location). 
%\todoam{Anche qui passa bene il messaggio del "multi-agent", meno quello del "partially observable" e "semi"}
%\todorp{La generality del setting e dell'approccio l'ho spostata nella conclusione.}

\section{Preliminaries}\label{sec:prelim}
%In this section, we provide the necessary background that will be employed in the successive sections.

\subsection{Configurable MDPs}
A Configurable Markov Decision Process (Conf-MDP) \cite{metelli2018configurable} is defined as a tuple $(\mathcal{S}, \mathcal{A}, r, \gamma, \mu, \mathcal{P}, \Pi)$, where $\mathcal{S}$ is the set of states, $\mathcal{A}$ is the set of actions, $r : \mathcal{S \times A} \rightarrow \mathbb{R}$ is the reward function specifying the reward $r(s,a)$ for taking action $a$ in state $s$, $\gamma \in (0,1)$ is the discount factor, $\mu \in \Delta(\mathcal{S})$\footnote{We denote with $\Delta(\mathcal{X})$ the set of probability distributions over a set $\mathcal{X}$.} is the distribution of the initial state, $\mathcal{P}$ and $\Pi$ are the model and policy spaces respectively. In particular, every $p \in \mathcal{P}$ is a transition function $p: \mathcal{S \times A} \rightarrow \Delta(\mathcal{S})$ that specifies a probability distribution $p(\cdot|s,a)$ over next state upon taking action $a$ in state $s$, and every $\pi \in \Pi$ is a policy $\pi: \mathcal{S} \rightarrow \Delta(\mathcal{A})$ specifying a probability distribution $\pi(\cdot|s)$ over actions for every state $s$. In Conf-MDPs, the goal is to find the optimal model-policy pair $(p^*, \pi^*) \in \mathcal{P} \times \Pi$ that maximizes the \emph{expected return}: $\mathcal{J}^{p,\pi} \coloneqq \expec{\sum_{t=0}^{+\infty} \gamma^t r(s_t, a_t)| \pi,p,s_0\sim \mu},$
where the expectation is taken w.r.t. the randomness of $\pi$, $p$, and $\mu$. 
This joint optimization is solved by two cooperating entities: the \emph{agent}, responsible for improving the policy $\pi$, and the \emph{configurator}, whose goal is to learn a configuration $p$.

\subsection{Semi-POMDPs}
A Semi-Partially Observable Markov Decision Process (Semi-POMDP) \cite{shi2020cooperative} is defined as a tuple $(\mathcal{D}, \mathcal{S}, \bm{\mathcal{A}}, p, r, \bm{\mathcal{O}}, o, \gamma, \mu)$, where $\mathcal{S}$, $\gamma$ and $\mu$ have the same meaning as before. $\mathcal{D}$ is the set of agents, $\bm{\mathcal{A}} = \mathcal{A}_1 \times \dots \times \mathcal{A}_{|\mathcal{D}|}$ is the set of joint actions the agents can perform,  $r : \mathcal{S} \times \bm{\mathcal{A}} \rightarrow \mathbb{R}$ is the reward function extended to joint actions, $\bm{\mathcal{O}}$ is the set of joint observations that are perceived by the agents, $o: \mathcal{S} \times \bm{\mathcal{A}} \rightarrow \Delta({\bm{\mathcal{O}}})$ is the observation function, that, for every state $s$ and joint action $\mathbf{a}$  provides a probability distribution $o(\cdot|s,\mathbf{a})$ over joint observations. The transition function $p: \mathcal{S} \times \bm{\mathcal{A}} \rightarrow \Delta(\mathcal{S} \times \mathbb{N})$ provides a probability distribution $p(\cdot,\cdot|s,\mathbf{a})$ over the next state and the time interval $k$ associated to the transition from the current state $s$ to the next one $s'$. 
%A joint policy $\bm{\pi}: \mathcal{S}
%and $p: \mathcal{S}_t \times \bm{\mathcal{A}} \rightarrow \Delta(S_{t+k})$ is a transition probability function between states given an action, and $k$ is a non-constant time interval from the current state to the next one. 
%\todoam{Il modello di transizione qui dovrebbe essere $p: \mathcal{S \times \bm{A}} \rightarrow \mathcal{S} \times \mathbb{N}$ che fornisce $(s',k)$ lo stato prossimo e un delta-step. Ne segue che la funzione obiettivo è:
%\[
%\mathcal{J}^{p,\bm{\pi}} = \expec{\sum_{i=0}^{+\infty} \gamma^{t_i} r(s_{t_i}, \mathbf{a}_{t_i})| \bm{\pi},p,s_0\sim \mu},
%\]
%where $t_i = \sum_{j=1}^{i} \tau_i$.
%}
A joint policy $\bm{\pi}$ maps a history of observations $\tau = (\mathbf{o}_0, \mathbf{a}_0, r_0, \mathbf{o}_1, \mathbf{a}_1, r_1, \dots)$ to a distribution over joint actions $\bm{\pi}(\cdot|\tau)$. The goal consists in finding an optimal policy $\bm{\pi}^*$ that maximizes the expected return: $\mathcal{J}^{p,\bm{\pi}} \coloneqq \expec{\sum_{i=0}^{+\infty} \gamma^{t_i} r(s_{t_i}, \mathbf{a}_{t_i})| \bm{\pi},p,s_0\sim \mu},$
where $t_0 = 0$ and $t_i = t_{i-1} + k_i$ for $i \ge 1$. 

%$\expec{\sum_{t=0}^{+\infty} \gamma^t r(s_t, \mathbf{a}_t)| \pi,p,s_0\sim \mu}$. 
%\todorp{Non sono sicuro si scriva cosi l'ottimizzazione per i Semi-POMDP.}
%\todoam{Ho cambiato la formulazione, fammi sapere}

\subsection{Empty Container Repositioning}
As highlighted in \cite{li2019cooperative}, the ECR problem can be modeled as a graph $\mathcal{G} \coloneqq (\mathrm{H}, \mathrm{V}, \mathrm{E})$, where $(\mathrm{H}, \mathrm{V}, \mathrm{E})$ are the set of harbor, vessels and routes respectively. 
More specifically, each harbor $h \in \mathrm{H}$ has a stock of empty containers of maximum capacity $C_h$. We denote with $C_h^t$ the number of containers available at day $t$. Each route $e \in \mathrm{E}$ is a directed cycle of consecutive harbors in $\mathrm{H}$, namely $e_k \coloneqq (h_1, \dots, h_{|e_k|})$, where $h_1 = h_{|e_k|}$. Routes can intersect with each other. Each vessel $v \in \mathrm{V}$ is associated with a maximum container capacity $C_v$ and a route $e \in \mathrm{E}$. We denote with $C_v^t$ the amount of empty space at day $t$ on vessel $v$. 
Finally, we denote with $u_v$ the speed function of vessel $v$. Given source and destination harbors $h_i$ and $h_j$, $u_v$ provides a probability distribution over the number of days required by $v$ to reach $h_j$ starting from $h_i$. 
Order of goods between ports are described by a stochastic function $q$. Given two ports $h_i, h_j \in \mathrm{H}$ and a day $t$, $q$ provides a probability distribution on the number of goods that are requested to be shipped from $h_i$ to $h_j$ at day $t$.
More specifically, $h_i$ can satisfy this demand using stock of empty containers at the previous day (i.e., $C_{h_i}^{t-1}$). Whenever this amount is not enough, a shortage of containers will happen. We denote with $L_{h}^t$ the total shortage on harbor $h$ at day $t$. 
When vessels arrive at harbors, the following happens:
\begin{itemize}
\item laden containers for that destination will be discharged to the port; after some days these containers will turn empty containers and will accumulate in the port's stock;
\item empty containers can be loaded/discharged on/from the vessel. These are the actions that the ECR policy is responsible for optimizing. 
\end{itemize}
As mentioned in \cite{li2019cooperative}, we remark that the behavior with which containers (both full and empty) are loaded/discharged to/from vessels is complex to be modeled. Indeed, this mechanism is subject to black-boxed and non-linear country regulations.

The goal is to find a policy that minimizes the total shortage, namely $\sum_{t, h \in \mathrm{H}} L_h^t$. A full mathematical model is available in Appendix A of \cite{li2019cooperative}. As one can easily verify, ECR can be formalized as a Semi-POMDP. More specifically, we note that agents (i.e., ports) are required to take repositioning actions when vessels arrive. The time that passes in between two subsequent vessel arrivals is non-constant and, in our case, stochastic (i.e., determined by function $u_v$).

\section{Related Works}\label{sec:literature}
\emph{Conf-MDPs} have been introduced in \cite{metelli2018configurable} for finite spaces, and extended in \cite{metelli2019reinforcement} for more complex continuous environments. In these seminal works, the agent is fully responsible for the configuration activity of the environment, which, in turn, results in an auxiliary task to optimize performance. As highlighted in \cite{metelli2018configurable}, this leads to a clear distinction between Conf-MDPs and multi-task learning \cite{vithayathil2020survey}. Indeed, in Conf-MDPs, the agent is not interested in learning and gathering experience samples in sub-optimal configuration; its interest is solely toward the optimal policy in the optimal environmental configuration. 
The configuration activity within the environment, as shown in more recent works \cite{metelli2019policy,ramponi2021learning}, can also be carried out by an external entity (i.e., configurator) whose goals can even be adversary w.r.t. the ones of the agent \cite{ramponi2021learning}. 
None of the previous methods, however, have been designed to handle the more complex multi-agent cooperative setting, in which numerous additional challenges are present (e.g., partial observability, highly dimensional states, intrinsic non-stationarity). \textit{Environment design} literature \cite{zhang2009general,ho2011multiagent}  is also related to configurable environments. However, substantial differences are present since these approaches assume that the configurator (interested party) has (partial) access to the agent's best response to a given environment. In addition, the agent's policies, given an environment, are fixed, and, consequently, the optimization process is not joint.

\emph{ECR problems} \cite{song2015empty} were historically solved using OR methods. However, the environment stochasticity, together with the non-linear and black-boxed constraints that are present in the problem, have led researchers to explore MARL solutions \cite{luo2018multi,li2019cooperative,shi2020cooperative}. Since the performance that a method can achieve in terms of satisfied demand in a given horizon strictly depends on the routes that the vessels will follow, \textit{network design} and \textit{fleet deployment} techniques have been proposed to jointly optimize the policy along with the fleet of sailing boats \cite{tran2015literature}. The difference between network design and fleet deployment is that, in network design methods, routes are generated together with assignments of vessel to routes; in FD, instead, the set of routes is fixed and pre-determined. The joint optimization problem has been investigated from an OR point of view in \cite{wang2017container}. However, in this more complex case, the pitfalls of OR methods are even amplified by the complexity of the problem. Furthermore, for large networks of vessels and ports, solving the joint problem with a mathematical programming formulation, leads to high computational requirements. For these reasons, hybrid approaches such as \cite{takano2011study} have been proposed. 
In \cite{takano2011study} the authors study the network design setting, and propose a method to generate paths for each of the vessels. Each network configuration is evaluated using the value of the objective function of the mixed-integer formulation of the ECR problem in that specific configuration. Then, they use Genetic Algorithms (GA) \cite{mirjalili2019genetic} to optimize for the configuration with the best objective function. This, however, inherits all the pitfalls of the OR approach. As our experiments will show, the plan can diverge from reality, leading to suboptimal solutions.

\section{The Configurable Semi-POMDP Framework}\label{sec:setting}
As we have seen, the Conf-MDP framework \cite{metelli2018configurable} models scenarios in which a configurator and a single agent cooperate to improve overall performance. In this section, we generalize the formulation to account for the peculiarities of the ECR+FD problem, i.e., the presence of multiple agents, the partial observability, and the semi-Markov property.

\begin{definition}\label{def:conf-semi-pomdp}
A \textup{Configurable Semi-POMDP} (Conf-Semi-POMDP) is a tuple $(\mathcal{D}, \mathcal{S}, \bm{\mathcal{A}}, r, \bm{\mathcal{O}}, o, \gamma, \mu, \mathcal{P}, \bm{\Pi})$, where $(\mathcal{D}, \mathcal{S}, \bm{\mathcal{A}}, r, \bm{\mathcal{O}}, o, \gamma, \mu)$ is a Semi-POMDP without transition function, and $\mathcal{P}$ and $\bm{\Pi}$ are the model and policy spaces.
\end{definition}

More precisely, $\bm{\Pi} = \Pi_1 \times \dots \times \Pi_{|\mathcal{D}|}$ is the set of history-dependent policies that the agents have access to (i.e., the set of ECR repositioning policies).
Thus, we can look at the novel Conf-Semi-POMDP framework as either (i) an extension of the Conf-MDP setting to semi-Markov, multi-agent, partially-observable environments or (ii) an extension of the Semi-POMDP to configurable environments in which we have no transition model $p$, that can indeed be altered as an effect of the environment configuration activity. We focus on the case where $\mathcal{P}$ is a parametric space of transition probability functions. This assumption, which is usual in Configurable MDPs \cite{metelli2018configurable}, nicely fits the ECR+FD domain, in which each $p \in \mathcal{P}$ encodes assignments of vessels to routes. More specifically, each configuration $p \in \mathcal{P}$ corresponds to $\{(v_i, e_i, h_i)\}_{i=1}^{|\mathrm{V}|}$, where each element $(v, e, h)$ encodes the fact that vessel $v$ follows route $e$ and starting from port $h$. We also enforce the constraint that $h$ must belong to $e$. 

The performance of a model-policy pair $(p, \pi) \in \mathcal{P} \times \Pi$ is defined via the expected return, as for Semi-POMDPs:
\begin{equation}\label{def:perf}
\mathcal{J}^{p, \bm{\pi}} \coloneqq \expec{\sum_{i=0}^{+\infty} \gamma^{t_i} r(s_{t_i}, \bm{a}_{t_i}) | s_0 \sim \mu, \bm{\pi}, p},
\end{equation}
where $t_0 = 0$ and $t_i = t_{i-1} + k_i$ for $i \ge 1$.  
%\todorp{Stessa nota di prima anche qui.}
%\todoam{Vedi nota sopra}
Thus, the goal, as for Conf-MDPs, consists of finding the optimal model-policy pair $(p^*, \bm{\pi}^*) \in \mathcal{P} \times \bm{\Pi}$ such that $\mathcal{J}^{p, \bm{\pi}}$ is maximized. We denote with $\bm{\pi}^*_p$ an optimal policy for a generic configuration $p \in \mathcal{P}$. For ease of notation, whenever it is clear from the context, we drop the specification of the environment in which a policy is run. For instance, $\mathcal{J}^{\bm{\pi}^*_p}$ measures the performance of pair $(p, \bm{\pi}^*_p)$. In our work, we consider the case in which a central entity (i.e., the shipping company) is responsible for optimizing/taking decisions on the adopted configuration $p$.

\section{Configure \& Conquer}\label{sec:method}
We now introduce Configure \& Conquer, our method to solve Conf-Semi-POMDPs. To appreciate its generality, we first present CC to solve a generic Conf-Semi-POMDP, and then discuss how it works in the ECR+FD domain. 

Imagine having an oracle that, given a model $p \in \mathcal{P}$, provides the performance index $\mathcal{J}^{\bm{\pi}^*_p}$ of the optimal policy $\bm{\pi}^*_p$ for that specific environment $p$. 
In this case, the original joint optimization problem described in Section \ref{sec:setting} reduces to: 
\begin{equation}\label{problem:simplified}
p^* \in \argmax_{p \in \mathcal{P}} \mathcal{J}^{\bm{\pi}^*_p}.
\end{equation}
In practice, however, we do not have access to such an oracle. Nevertheless, given a configuration $p$, it is possible to train an algorithm $\mathscr{A}_c$ of our choice to learn an approximation of the optimal policy $\tilde{\bm{\pi}}^*_{p}$, and, consequently, $\mathcal{J}^{\tilde{\bm{\pi}}^*_{p}}$. In particular, there might exist sample efficient (yet suboptimal) algorithms that can be used to obtain such approximations. In that case, we can leverage these methods to optimize the empirical version of the objective function $\mathcal{J}^{\tilde{\bm{\pi}}^*_p}$, where the expectation in $\mathcal{J}^{p,\bm{\pi}}$ is estimated with trajectories collected within model $p$ using $\tilde{\bm{\pi}}^*_p$.
We can notice that, with this new formulation, the contribution of the configurator to the optimal solution is completely decoupled from the problem of finding the optimal agents' policy. Indeed, when the approximation of $p^*$ is found (i.e., \textit{configure step}), \algo optimizes the agents' policy in the tuned environment with a more complex and expensive algorithm $\mathscr{A}_*$ (i.e., \textit{conquer step}), which aims at obtaining better approximations of $\bm{\pi}^*_{p^*}$. For this reason, \algo is a two-stage optimization algorithm. The general pseudo-code is reported in Algorithm \ref{algo:general}.

With respect to the ECR+FD setting, Algorithm \ref{algo:general} evaluates a given assignment $p$ of vessels to routes exploiting a cheaper algorithm $\mathscr{A}_c$ to train a cooperative ECR policy in $p$. Once this is done, $\mathscr{A}_*$ is used to train the final ECR policy that will be deployed in the approximation of optimal fleet $\tilde{p}_*$. We now discuss how to choose $\mathscr{A}_*$ and $\mathscr{A}_c$.

%\todoam{Non diciamo più niente riguardo alle garanzie teoriche. Non dico di inserirle, ma magari fare riferimento al fatto che un punto di forza di CC consiste nel fatto che possiamo sfruttare anche approssimazioni poco accurate purchè l'ordinamento sia mantenuto, secondo me puo' dare valore alla proposta.}

\subsection{Choosing $\mathscr{A}_c$ and $\mathscr{A}_*$. }
The choice of algorithms for computing approximations of the optimal policy in some configuration $p$ depends on the specific
problem at hand. For what concerns $\mathscr{A}_c$, there is a trade-off between computational/sample efficiency and performance. Indeed, since $\mathscr{A}_c$ is used to train the
configurator, the ideal method should be fast to compute and provide good approximations of $\bm{\pi}^*_p$. The main issue is that the faster the method, the more configurations we can evaluate in a reasonable amount of time so to find $\tilde{p}^*$. However, this usually comes at the cost of precision, which might impact the optimization landscape of $\mathcal{J}^{\tilde{\bm{\pi}}^*_p}$. For these reasons, depending on the problem, one might use heuristics, mathematical programming, experts, reinforcement learning agents trained on limited data and so on. On the other hand, the choice of $\mathscr{A}_*$ is more critical for the final performance of the system since it is responsible for computing agents' policy that will be actually deployed in $\tilde{p}^*$. In this sense, the optimal choice for $\mathscr{A}_*$ is the state-of-the-art for the considered industrial setting. 

In our experiments, we analyze the performance of \algo varying $\mathscr{A}_c$ and $\mathscr{A}_*$ among traditional methods usually employed in ECR domains. More specifically, we consider algorithms that ranges from simple heuristics, to the more complex OR \cite{li2019cooperative} and MARL approaches \cite{li2019cooperative,shi2020cooperative}. 

\begin{algorithm}[t]
\caption{Configure \& Conquer (CC).} \label{algo:general}
\begin{algorithmic}[1]
\REQUIRE Algorithms $\mathscr{A}_*$, $\mathscr{A}_c$, model and policy spaces $\mathcal{P}$, $\bm{\Pi}$
\vspace{-0.1cm}
\STATE{Solve $\tilde{p}^* \in \argmax_{p \in \mathcal{P}} \mathcal{J}^{\tilde{\bm{\pi}}^*_p}$ using $\mathscr{A}_c$ to estimate $\tilde{\bm{\pi}}^*_p$}
\STATE{ Solve $\tilde{\bm{\pi}}^*_{\tilde{p}^*} \in \argmax_{\bm{\pi} \in \bm{\Pi}} \mathcal{J}^{\tilde{p}^*,\bm{\pi}}$ using $\mathscr{A}_*$}
\end{algorithmic}
\end{algorithm}

% devo anche dire la cosa dell'hidden assumption e della sua discussione teorica... forse sarebbe meglio introdurlo il piu tardi possibile. Probabilmente
% ne vale la pena. Forse addirittura mettere addirittura tutto quanto dentro il teorema e anche dopo la spiegazione dei GA e di RL per risolvere il problema.

\subsection{Configurator Optimization}\label{sec:conf-opt}
We have seen how CC decouples the joint optimization process into two subsequent stages: configure and conquer. More specifically, once the configuration step is over, the conquer step consists in simply applying any algorithm of choice in the tuned environmental setting. For this reason, the crucial step of CC consists in finding an approximation of $p^*$. We now provide an in-depth description of how one can use Reinforcement Learning (RL) to solve this issue.

Suppose w.l.o.g. that $D$ is the dimension of the parametric space $\mathcal{P}$ (i.e., the number of parameters required to define every $p \in \mathcal{P}$). At this point, it is possible to construct an MDP whose sequence of actions will define a configuration $p \in \mathcal{P}$. More specifically, at each timestep the agent chooses a dimension-value pair $(d, z)$ that will assign the value $z$ for dimension $d$. At time $t$, the state contains information about the previously selected pairs $\{(d_i,z_i)\}_{i=1}^{t-1}$, and dummy values for the missing configuration parameters. After $D$ timestamps, an entire configuration $p$ is produced and the agent will receive a reward according to $\mathcal{J}^{\tilde{\bm{\pi}}_p}$, where $\tilde{\bm{\pi}}^*_p$ is the approximation of the optimal policy of algorithm $\mathscr{A}_c$ in model $p$. For all previous timestamps, the reward is fixed at $0$ for any action. Given this formulation, it is easy to show that the expected discounted reward of the fixed initial state (i.e., no value-dimension assignment done) is proportional to $\E_{p \sim \nu_c} \left[ \mathcal{J}^{\tilde{\bm{\pi}}_p} \right]$, where $\nu_c$ is the single-agent configurator policy. At this point, one can use any RL algorithm to train the configurator to solve this MDP. In our experiments, we model $\nu_c$ with neural networks, and rely on PPO \cite{schulman2017proximal} for its optimization.

We notice that, in the ECR+FD setting, the configurator, at each step, needs to assign a vessel to a route and an initial port; i.e., it selects a triplet $(v, e, h)$. It follows that, with the most naive implementation of $\nu_c$, the action space would be given by the cartesian product of $H \times V \times E$, which, in practice, can be quite large, thus making the learning process hard, unstable, and inefficient.\footnote{In our experiments, we consider real-world instances of $46$ ships, $22$ ports and $13$ routes. This means that the action space would have dimension $13156$.} Moreover, we expect a good configurator to be able to exploit the inner structure within the parametric space $\mathcal{P}$, such as similarities between configurations.  Immagine two distinct configurations $p_1,p_2 \in \mathcal{P}$ in which the only difference is that a given vessel $\overline{v} \in V$ is assigned to the same route $\overline{e} \in E$ but two different initial ports $h_1$ and $h_2$ in $\overline{e}$. Clearly, we can expect the rewards $\mathcal{J}^{\tilde{\bm{\pi}}_p}$ that the configurator will obtain for the two configurations to be very similar. 
Therefore, to make training efficient and effective, we rely on the following more complex architecture for $\nu_c$. First of all, we reduce significantly the action space size using \emph{autoregressive} policies \cite{metz2017discrete}. More specifically, each action (i.e., triplet $(v,e,h)$), is split into three components (i.e., $v$, $e$ and $h$) that will be sequentially picked one before the next. We first pick the route $e$, then given the route, we pick the port $h$, and given the route and the port we pick the vessel $v$. By doing so the size of the action space is reduced from $|H||V||E|$ to $|H| + |V| + |E|$. Moreover, to exploit similarity between configurations, we enlarge the state of the agent at timestep $t$ with features of the current uncompleted configuration (e.g., number of vessels assigned to each route, total capacity of the vessels assigned to each route). These features are processed by a neural network $f$ to create an embedding of the current state of the agent. This embedding is used to select the first sub-action (i.e., the route), and it is concatenated to previous selected sub-actions to compute the next sub-actions. Moreover, to further exploit structure the parametric space $\mathcal{P}$, any unfeasible action (e.g, a port that does not exist in a route) is masked.

\section{Experiments}\label{sec:exp}

%\begin{figure}[!t]
%\centering
%\includegraphics[width=3.5in]{globaltrade}
%\caption{World-trade cooperative marine transportation network. Line with the same color represents a route, while points on the map represents a port. In our experiments, we adopt similar topologies.}
%\label{fig_wwt}
%\end{figure}

\subsection{Experimental Setup}
Similar to previous studies \cite{li2019cooperative,shi2020cooperative}, our experiments aims at testing our approach in scenarios that mimic dimensions and behaviors of international transportation companies. To this end, we rely on a patched version of the MARO simulator \cite{MARO_MSRA}. More specifically, we test our method on two different topologies $WWT_1$ and $WWT_2$. $WWT_1$ is composed of $46$ vessels, $22$ ports, and $13$ routes (as in \cite{shi2020cooperative}); in $WWT_2$, the number of ports and vessels is the same, but the number of routes is $6$. Moreover, in $WWT_1$, we consider an optimization horizon of $400$ days, while in $WWT_2$, $200$ days are considered. For both problems, as in previous works \cite{shi2020cooperative}, order distributions have complex trigonometric shapes with multiple periods.

Given this real-world inspired setup, our experiments aim at answering the following questions: (i) Can \algo find better configurations in which the agents operate? (ii) How does the performance of $\mathscr{A}_c$ and $\mathscr{A}_*$ impact the final results? (iii) How does \algo compare to existing algorithms? To this end, we conduct an extensive empirical study of \algo in $WWT_1$ and $WWT_2$, picking as $\mathscr{A}_c$ and $\mathscr{A}_*$ the following algorithms:
\begin{itemize}
\item \textbf{Random policy (Rand).} A random repositioning action is taken every-time a vessel arrives at a certain port. 
\item \textbf{Heuristic policy (Heur).} This is a stochastic intuitive heuristic that we propose to solve ECR problems. If a port is an exporting one (i.e., it exports much more goods w.r.t. the ones that it imports), then, whenever a vessel arrives, we randomly discharge at least parts of the empty containers that it carries. If a port is an importing one (i.e., it imports much more goods than the ones it exports), instead, we randomly load empty containers on the vessel.
\item \textbf{Operation Research (OR).} Noisy estimates of future orders and vessel arrivals are used at the beginning of the interaction with the environment to compute a plan by solving the mathematical formulation of the problem (see Appendix in \cite{li2019cooperative}).
\item \textbf{Operation Resarch methods with iterative plan (OR(I)).} Noisy estimates of future orders and vessel arrivals are used to solve the mathematical formulation of the ECR problem. The plan is computed for a long horizon but executed only for a short window. Once the window expires, a new plan is recomputed using the current state of the environment, so to prevent the plan to diverge from reality \cite{long2012sample,li2019cooperative}.
\item \textbf{MARL system}. The application of MARL techniques to ECR problems has been studied in several works \cite{luo2018multi,li2019cooperative,shi2020cooperative}. In our experiments, we use a variant of \cite{shi2020cooperative}.
\end{itemize}
We select a subset of combinations of $\mathscr{A}_c$ and $\mathscr{A}_*$ that highlight our contributions and that successfully answer to the previous questions. The notation that will be used for \algo is CC-$\mathscr{A}_c$-$\mathscr{A}_*$.

Moreover, we compare \algo with the following baselines:
\begin{itemize}
\item \textbf{LS-NET \cite{takano2011study}.} LS-NET \cite{takano2011study} was originally proposed to tackle the joint problem of \textit{network design} and ECR, however, the extension to the ECR+FD is direct. We define elements of a GA population so that each element describes a configuration $p \in \mathcal{P}$ (i.e., assignments of vessels to routes and initial ports), and we evaluate each of the elements in the population using as fitness function the value of the objective function of the OR formulation of the problem. Once the method has reached convergence, OR(I) is used to evaluate the performance of the approximation of the optimal configuration. 
\item \textbf{Genetic Algorithm - Joint (GA joint).} GAs are used to jointly optimize the configuration and the agents policy. The agents' policy is represented as a matrix in which cell $(i,j)$ specifies the repositioning $j$-th action of the $i$-th vessel. Configurations are represented as assignments of vessels to routes. An element in the GA population is, thus, a concatenation of a policy with a configuration.
\item \textbf{Random Configuration and OR (RandomConf-OR(I)).} Configurations are generated at random; OR(I) computes the policy on these sampled configurations.
\end{itemize}

\subsection{Results}
Table \ref{tab:results-single-task} reports mean and $95\%$ confidence intervals (5 runs) of the percentage of satisfied domand of different algorithms on $WTT_1$ and $WTT_2$. We highlight in bold the highest performance reached in each domain. 

\begin{table}[!t]
\caption{ECR+FD results (5 runs, mean $\pm$ $95\%$ c.i.).\label{tab:results-single-task}}
\centering
\begin{tabular}{l|c|c}
\hline
Algorithm       & $WTT_1$   & $WTT_2$  \\
\hline
CC-Heur-Heur   			& $86.21 \pm 0.15$ 	&  $78.64 \pm 0.27$		\\
CC-Heur-OR(I)     		& $90.93 \pm 0.34$ 	&  $90.58 \pm 0.39$		\\
CC-Rand-Rand   			& $77.72 \pm 0.26$  	&  $42.69 \pm 0.11$		\\
CC-Rand-OR(I)     		& $87.93 \pm 0.12$  	&  $87.15 \pm 0.35$		\\
CC-OR-OR 			& $88.38 \pm 0.23$ 	&  $86.15 \pm 0.23$		\\
CC-OR-MARL  			& $88.77 \pm 0.41$  	&  $\mathbf{94.88 \pm 0.60}$		\\
CC-OR-OR(I)    		& $90.95 \pm 0.12$	&  $93.03 \pm 0.60$ 		\\
CC-OR(I)-OR(I) 			& $\mathbf{92.28 \pm 0.35}$	&  $\mathbf{94.52 \pm 0.47}$ 		\\
CC-OR(I)-MARL  	   		& $88.85 \pm 0.21$	&  $\mathbf{95.35 \pm 0.88}$ 		\\
CC-OR(I)-Rand  			& $74.16 \pm 1.66$	&  $39.15 \pm 5.45$		\\ \hdashline[4pt/2pt]
GA joint    				& $86.32 \pm 0.18$ 	&  $83.73 \pm 0.90$		\\
LS-NET         			& $88.21 \pm 0.54$  	&  $84.66 \pm 1.70$	 	\\
RandomConf-OR(I)        	& $77.42 \pm 0.18$  	&  $68.49 \pm 0.56$		\\
\hline
\end{tabular}
\end{table}

First of all, as long as we choose good algorithms for $\mathscr{A}_*$ (i.e., OR, OR(I), MARL), we can notice that \algo is able to find configurations in which the agents operate that are better than random. This is confirmed by the fact that a good method on random configurations (i.e., RandomConf-OR(I)) reaches lower performance w.r.t. cases in which configurations have been tuned using \algo. In this sense, the choice of $\mathscr{A}_*$ is the one that most impact the final performance of our two-stage optimization algorithm. This is expected; indeed, even though we find the optimal configuration, if our agents ignore how to behave (see CC-OR(I)-Rand), the performance will be highly sub-optimal (even worse than using a good method on random configurations such as RandConfig-OR(I) does). The more sophisticated $\mathscr{A}_*$ is (e.g., MARL and OR(I)) the better the performance we can obtain. As for the choice of $\mathscr{A}_c$, instead, most surprisingly we notice that its choice does not affect performance significantly. Indeed, even when the reward for the configurator is computed using random policies, one can still significantly improve the performance. In particular, we notice that in $WTT_2$, even if the random policy is highly suboptimal (see CC-Rand-Rand; $42.69\%$), \algo is still able to find configurations that lead to significantly good performance (see CC-Rand-OR(I); $87.15\%$). Finally, we see that GA joint and LS-NET  underperform w.r.t. all the non-ablation versions of \algo. For what concerns GA joint, we conjecture that its sub-optimality arises from the difficulty of the joint optimization process. LS-NET, on the other hand, computes the plan beforehand and optimizes the configuration relying on the value of an objective that might not be respected in practice. Indeed, if the value of the objective function of the mathematical formulation (i.e., the fitness function used in the GA optimization steps) diverges from reality, then LS-NET optimizes toward sub-optimal configurations.

\section{Conclusions}
In this work, we have studied the ECR+FD setting under the novel perspective of Configurable Semi-POMDPs. We have modeled the fleet deployment problem as configurations of the environment that can be tuned to increase the overall performance of the multi-agent system. In particular, we focused on the case in which decisions on the fleet of vessels are taken by a central entity (i.e., the shipping company) that cooperates with the agents to minimize the total shortage of containers. We proposed a novel two-stage optimization algorithm (\algo) that, as our experiments show, successfully solve the joint ECR+FD optimization problem in large and real-world inspired problem instances, outperforming competitive baselines. 

We remark the generality of the proposed approach. Indeed, a broad number of multi-agent environments have the possibility to be configured to maximize the performance of the system. In this sense, \algo represents a viable option that can successfully scale to large and complex domains. Our work represents a further step in the literature of Configurable MDPs, that, so far, have  managed to solve problem of much smaller dimensions only. We also notice that \algo, can be directly applied as-is in single-agent problems as well.  

As future works, we plan to apply the Conf-Semi-POMDPs framework to the more complex ECR+network design setting. In this case, the set of routes is not pre-determined but paths for each of the vessels are computed by the configurator.

%Our work opens up several exciting research directions. From the point of the configurable environment literature, we could try to extend \algo to reuse the data collected during the first stage in the second one. In this sense, the off-policy multi-agent RL method \cite{li2019cooperative} may represent a good choice for $\mathscr{A}_*$. In addition, one could reason about reusing data collected in similar configurations to further improve the sample efficiency of the method. Other directions for future work are: the design of algorithms with strong theoretical guarantees and the study of the multi-agent configurable setting in competitive domains. For what concerns resource optimization in marine transportation, instead, we could try to apply Conf-Semi-POMDPs to the more complex ECR-network design case (i.e., the set of routes is not pre-determined but paths for each of the vessels are computed by the configurator).

%\section*{Acknowledgments}
%This should be a simple paragraph before the References to thank those individuals and institutions who have supported your work on this article.

%\section*{Multi-agent Reinforcement Learning for ECR}
%You can choose not to have a title for an appendix if you want by leaving the argument blank
%\section*{}


{
\bibliographystyle{IEEEtran}
\bibliography{biblio}

\begin{thebibliography}{10}
\providecommand{\url}[1]{#1}
\csname url@samestyle\endcsname
\providecommand{\newblock}{\relax}
\providecommand{\bibinfo}[2]{#2}
\providecommand{\BIBentrySTDinterwordspacing}{\spaceskip=0pt\relax}
\providecommand{\BIBentryALTinterwordstretchfactor}{4}
\providecommand{\BIBentryALTinterwordspacing}{\spaceskip=\fontdimen2\font plus
\BIBentryALTinterwordstretchfactor\fontdimen3\font minus
  \fontdimen4\font\relax}
\providecommand{\BIBforeignlanguage}[2]{{%
\expandafter\ifx\csname l@#1\endcsname\relax
\typeout{** WARNING: IEEEtran.bst: No hyphenation pattern has been}%
\typeout{** loaded for the language `#1'. Using the pattern for}%
\typeout{** the default language instead.}%
\else
\language=\csname l@#1\endcsname
\fi
#2}}
\providecommand{\BIBdecl}{\relax}
\BIBdecl

\bibitem{long2012sample}
Y.~Long, L.~H. Lee, and E.~P. Chew, ``The sample average approximation method
  for empty container repositioning with uncertainties,'' \emph{EJOR}, 2012.

\bibitem{li2019cooperative}
X.~Li, J.~Zhang, J.~Bian, Y.~Tong, and T.-Y. Liu, ``A cooperative multi-agent
  reinforcement learning framework for resource balancing in complex logistics
  network,'' in \emph{AAMAS}, 2019, pp. 980--988.

\bibitem{song2015empty}
D.-P. Song and J.-X. Dong, ``Empty container repositioning,'' \emph{Handbook of
  ocean container transport logistics}, pp. 163--208, 2015.

\bibitem{tran2015literature}
N.~K. Tran and H.-D. Haasis, ``Literature survey of network optimization in
  container liner shipping,'' \emph{Flexible Services and Manufacturing
  Journal}, vol.~27, no.~2, pp. 139--179, 2015.

\bibitem{wang2017container}
S.~Wang and Q.~Meng, ``Container liner fleet deployment: a systematic
  overview,'' \emph{TR\_C}, 2017.

\bibitem{shi2020cooperative}
W.~Shi, X.~Wei, J.~Zhang, X.~Ni, A.~Jiang, J.~Bian, and T.-Y. Liu,
  ``Cooperative policy learning with pre-trained heterogeneous observation
  representations,'' in \emph{AAMAS}, 2021, pp. 1191--1199.

\bibitem{metelli2018configurable}
A.~M. Metelli, M.~Mutti, and M.~Restelli, ``Configurable markov decision
  processes,'' in \emph{ICML}, 2018, pp. 3491--3500.

\bibitem{metelli2019policy}
A.~M. Metelli, G.~Manneschi, and M.~Restelli, ``Policy space identification in
  configurable environments,'' \emph{Machine Learning}, pp. 1--53, 2021.

\bibitem{metelli2019reinforcement}
A.~M. Metelli, E.~Ghelfi, and M.~Restelli, ``Reinforcement learning in
  configurable continuous environments,'' in \emph{ICML}, 2019.

\bibitem{puterman}
M.~L. Puterman, \emph{Markov Decision Processes: Discrete Stochastic Dynamic
  Programming}, 1st~ed., USA, 1994.

\bibitem{ramponi2021learning}
G.~Ramponi, A.~M. Metelli, A.~Concetti, and M.~Restelli, ``Learning in
  non-cooperative configurable {M}arkov decision processes,'' \emph{NeurIPS},
  vol.~34, 2021.

\bibitem{zhang2021multi}
K.~Zhang, Z.~Yang, and T.~Ba{\c{s}}ar, ``Multi-agent reinforcement learning: A
  selective overview of theories and algorithms,'' \emph{Handbook of
  Reinforcement Learning and Control}, pp. 321--384, 2021.

\bibitem{vithayathil2020survey}
N.~Vithayathil~Varghese and Q.~H. Mahmoud, ``A survey of multi-task deep
  reinforcement learning,'' \emph{Electronics}, 2020.

\bibitem{zhang2009general}
H.~Zhang, Y.~Chen, and D.~C. Parkes, ``A general approach to environment design
  with one agent,'' in \emph{IJCAI}, 2009.

\bibitem{ho2011multiagent}
C.-J. Ho, Y.-L. Kuo, and J.~Y.-j. Hsu, ``Multiagent environment design in human
  computation,'' in \emph{AAMAS}, 2011, pp. 1279--1280.

\bibitem{luo2018multi}
Q.~Luo and X.~Huang, ``Multi-agent reinforcement learning for empty container
  repositioning,'' in \emph{ICSESS}.\hskip 1em plus 0.5em minus 0.4em\relax
  IEEE, 2018.

\bibitem{takano2011study}
K.~Takano and M.~Arai, ``Study on a liner shipping network design considering
  empty container reposition,'' \emph{JASNAOE}, 2011.

\bibitem{mirjalili2019genetic}
S.~Mirjalili, ``Genetic algorithm,'' in \emph{Evolutionary algorithms and
  neural networks}.\hskip 1em plus 0.5em minus 0.4em\relax Springer, 2019, pp.
  43--55.

\bibitem{schulman2017proximal}
J.~Schulman, F.~Wolski, P.~Dhariwal, A.~Radford, and O.~Klimov, ``Proximal
  policy optimization algorithms,'' \emph{arXiv preprint arXiv:1707.06347},
  2017.

\bibitem{metz2017discrete}
L.~Metz, J.~Ibarz, N.~Jaitly, and J.~Davidson, ``Discrete sequential prediction
  of continuous actions for deep rl,'' \emph{arXiv preprint arXiv:1705.05035},
  2017.

\bibitem{MARO_MSRA}
\BIBentryALTinterwordspacing
``Maro: A multi-agent resource optimization platform,'' 2020. [Online].
  Available: \url{https://github.com/microsoft/maro}
\BIBentrySTDinterwordspacing

\end{thebibliography}
}

%\newpage


%\begin{IEEEbiography}[{\includegraphics[width=1in,height=1.25in,clip,keepaspectratio]{riccardo}}]{Riccardo Poiani}
%Riccardo Poiani received the bachelor's degree in computer science from Politecnico di Milano in 2015, and the master's degree in computer science from Politecnico di Milano in 2021. Once graduated he joined InstaDeep (Paris, France), where he started working on multi-agent systems for marine resource optimization. From November 2021, he is pursuing the PhD degree at Politecnico di Milano, advised by Professor Marcello Restelli. His current research interest is Reinforcement Learning, with a focus on non-stationary learning, meta-learning and transfer learning.
%\end{IEEEbiography}


%\vfill

\end{document}